# Off-Line Arabic Handwriting Character Recognition Using Word Segmentation

Manal A. Abdullah, Lulwah M. Al-Harigy, and Hanadi H. Al-Fraidi

**Abstract**—The ultimate aim of handwriting recognition is to make computers able to read and/or authenticate human written texts, with a performance comparable to or even better than that of humans. Reading means that the computer is given a piece of handwriting and it provides the electronic transcription of that (e.g. in ASCII format). Two types of handwriting: on-line and off-line. The most important purpose of off-line handwriting recognition is in protection systems and authentication. Arabic Handwriting scripts are much more complicated in comparison to Latin scripts. This paper introduces a simple and novel methodology to authenticate Arabic handwriting characters. Reaching our aim, we built our own character database. The research methodology depends on two stages: The first is character extraction where preprocessing the word and then apply segmentation process to obtain the character. The second is the character recognition by matching the characters comprising the word with the letters in the database. Our results ensure character recognition with 81%. We eliminate FAR by using similarity percent between 45-55%. Our research is coded using MATLAB.

——————————— ◆ ———————————

## 1 INTRODUCTION

Pattern recognition is the scientific discipline whose goal is the classification of objects into number of categories or classes. Depending on the application, these objects can be images or signal waveforms or any type of measurements that need to be classified [1]. In pattern recognition field, languages' recognition is considered as one of the most complicated problem in Artificial Intelligent field.

Handwriting recognition is one of the very challenging problems. An overview of word handwriting recognition

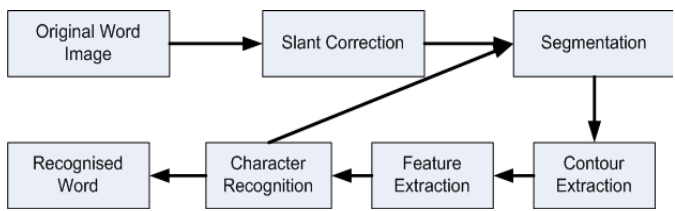

Figure1: An Overview of handwriting word recognition process [2].

process is shown in figure1 [2]:

There are two types of handwriting recognition: off-line recognition and on-lsine recognition. In off-line recognition, only the image of the handwriting is available for the computer, while in the on-line case temporal information such as pen tip coordinates as a function of time is also available.
Typical data acquisition devices for off-line and on-line recognition are scanners and digitizing tablets, respectively. Due to the lack of temporal information, off-line handwriting recognition is considered more difficult than on-line. Furthermore, it is also clear that the off-line case is the one that corresponds to the conventional reading task performed by humans [3].

The typed Latin, Chinese or Japanese scripts are widely used around the world. Their characters are separated from one another which make their Optical Character Recognition OCR techniques easier to develop. These are the reasons why OCR systems for these scripts are well developed and most commercial available OCR systems recognize any of these three scripts. Arabic is another popular script. It is estimated that there are over one billion Arabic script users. However, because of the technical difficulties induced by the cursive nature of the Arabic script, its OCR techniques have not been well developed yet. If OCR systems are available for Arabic characters, they will be very useful and have a great commercial value [4].The Arabic characters unlike the Latin characters are very similar to each other. For example the only difference between the letters ح, خ and ج is the dot position; this is also true for the letters ب, ت and ث and so on. The discrimination of such characters can be difficult which requires sensitive classifiers. The similarity of the characters can cause further problems with classification when noise is added to these characters [5].

## 2 LITERATURE REVIEW

Since the end of 1980s, the very successful use of Hidden Markov Models HMMs in speech recognition has led many researchers to apply them to various problems in the field of handwriting recognition such as character

————————————————
- *M.A. Abdullah is with the Department of Computer Science, Faculty of Computing and Information Technology FCIT, King Abdulaziz University KAU, Jeddah.*
- *L.M. Al-Harigy is with the Department of Computer Science, Faculty of Computing and Information Technology FCIT, King Abdulaziz University KAU, Jeddah.*
- *H.H. Al-Fraidi is with the Department of Computer Science, Faculty of Computing and Information Technology FCIT, King Abdulaziz University KAU, Jeddah, KSA* recognition, offline word recognition, and signature verification and identification. These HMM frameworks can be distinguished from each other by the state meaning, the modeled units (stroke, character, word, etc.), the unit model topology (ergodic or left-to-right), the HMM type (discrete or continuous), the HMM dimensionality (one-dimensional, planar, bidimensional, or random fields), the state duration modeling type (implicit or explicit), and the modeling level (morphological, lexical, syntactical, etc.). The results obtained in [6] were very promising and had shown that the explicit state duration modeling within HMM framework can improve the recognition rate significantly. Moreover, continuous distributions (i.e., Gamma and Gauss) of state duration were more suitable than discrete ones (i.e., Poisson) for Arabic handwriting modeling, and the nonuniform segmentation scheme was more recommended. The main drawback of discrete HMMs was the imperfect observation probability estimation.

Gillies [7] used an implicit segmentation-based HMM for cursive word recognition. First, a label is given to each pixel in the image according to its membership in strokes, holes, and concavities. Then, the image is transformed into a sequence of symbols by vector quantization of each pixel column. Each letter is modeled by a different discrete HMM whose parameters are estimated from hand segmented data. The Viterbi algorithm is used for recognition and it allows an implicit segmentation of the word into letters by a by-product of the word matching.

Mohamed and Gader [8] used continuous HMMs to segmentation-free modeling of handwritten words in which the observations are based on the location of black-white and white-black transitions on each image column. They designed a 12-state left-to-right HMM for each character.

Somaya et al [9] presented a complete scheme for totally unconstrained Arabic handwritten word recognition based on a Model discriminate HMM is presented. They proposed and discussed a complete system able to classify Arabic-Handwritten words of one hundred different. The system first attempts to remove some of variation in the images. Next, the system codes the skeleton and edge of the word so that feature information about the lines in the skeleton is extracted. Then they used a classification process based on the HMM approach. The output is a word in the dictionary.

Sabri A. and Sameh M. [10] described a technique for automatic recognition of off-line handwritten Arabic (Indian) numerals using Support Vector Machines (SVM) and Hidden Markov Models. In addition, SVM classifier results are compared to those of the HMM classifier. The SVM and HMM classifiers were trained with 75% of the data and tested with the remaining data. Other divisions of data for training and testing were performed and resulted in comparable performance. The achieved average recognition rates were 99.83% and 99.00% using, respectively, the SVM and HMM classifiers. SVM recognition rates proved to be better for all digits. Comparison at the writer's level (Writers 34 to 44) showed that SVM results outperformed HMM results for all tested writers. The presented technique, using the powerful set of features and the SVM classifier, proved to be effective in the recognition of independent writer Arabic (Indian) numerals and was shown to be superior to the HMM classifier.

## 3 RESEARCH METHODOLOGY

Our research uses the concept of word segmentation where the characters are recognized using preprocessed character database. Figure 2 shows system architecture for Arabic Handwriting recognition. It is about how to recognize Arabic characters from a word. The input image goes through the steps of preprocessing, segmentation and recognition.

### 3.1 Word Extraction

This stage aims to remove the white spaces around each word and extract the effective image size by moving through the black pixels and when it finds white space more than 100 pixels that means the word is finished and it starts another word.

### 3.2 Preprocessing

The aim of preprocessing stage is the removal of all elements in the word image that are not useful for recognition process [11]. It includes:
1. Convert RGB image or colormap to grayscale (rgb2gray)
2. Define a Crop function charcrop (crop the white area around the word).

### 3.3 Segmentation

The word is divided into segments with specific width. The width value depends on the width of the word. After a lot of trails based on the letter width we choose the segment width to start with 35 pixels where we find this width matches the most thin characters such as "أ" . If the character is not recognized, the program resizes the segment width by some factor and widens the segment and repeats the recognition process again. We conduct many trails to decide the widening factor which gives the highest recognition rate. We used three words: "زرع , جامعة , شمس" for test.

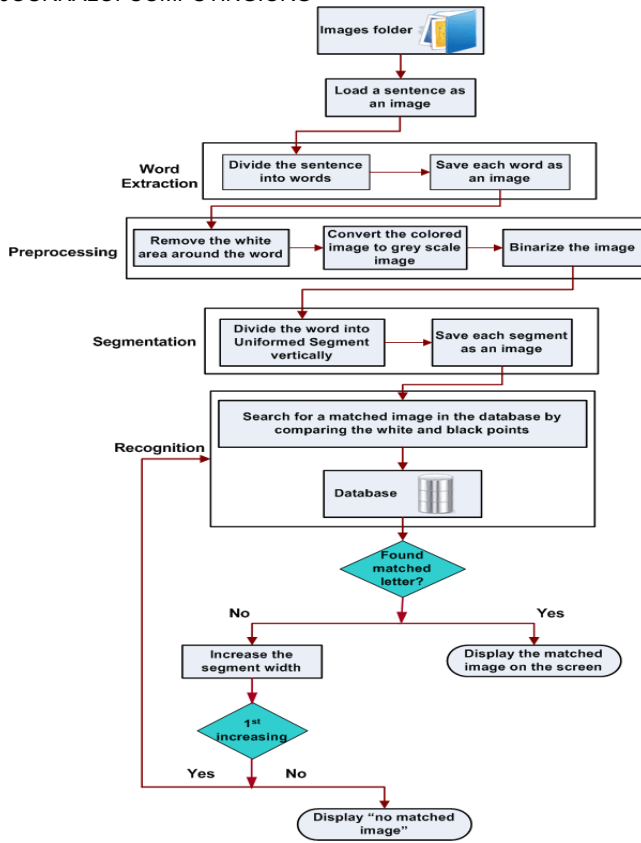

Figure2: Arabic HW Recognition process flow diagram.

### 3.4 Character Recognition

The algorithm used in our research is detecting black and white points in the two pictures and comparing the white points (edge points) between them by finding the matched data. Then the algorithm calculates the percentage of matching by dividing the matched data over the total data. We decided that if the similarity percentage (matched percentage) between the two images is 45% or more, then we accept the result and display the matched image from the database.

### 3.5 Database Description

We search for an available DB that can be used for character recognition. The only available DB is **IFN/ENIT** [1] which consists of Tunisian names and it is designed for Arabic words recognition. That is why we decided to create our own database.

Our DB consists of all Arabic letters by scan handwritten Arabic alphabets (written by two writers) as images. The alphabets written in different position shapes (Isolated, Initial, Medial, and Final) as illustrated in table 1. Each letter represented in four to ten shapes saved as JPG format with the same size (38 pixels width × 50 pixels height) and 100pixels/inch resolution. Table 1 shows Arabic characters DB.

TABLE 1
THE ARABIC ALPHABET IN DIFFERENT POSITIONS

| Name | Isolated | Initial | Medial | Final |
|---|---|---|---|---|
| Alif | أ | ا | ـا | ـا |
| Baa | ب | بـ | ـبـ | ـب |
| Taa | ت | تـ | ـتـ | ـت |
| Thaa | ث | ثـ | ـثـ | ـث |
| Jeem | ج | جـ | ـجـ | ـج |
| 7aa | ح | حـ | ـحـ | ـح |
| 7haa | خ | خـ | ـخـ | ـخ |
| Daa | د | د | ـد | ـد |
| Thal | ذ | ذ | ـذ | ـذ |
| Raa | ر | ر | ـر | ـر |
| Zaa | ز | ز | ـز | ـز |
| Seen | س | سـ | ـسـ | ـس |
| Sheen | ش | شـ | ـشـ | ـش |
| Saad | ص | صـ | ـصـ | ـص |
| Dhaad | ض | ضـ | ـضـ | ـض |
| 6aa | ط | ط | ـط | ـط |
| 6haa | ظ | ظ | ـظ | ـظ |
| Aeen | ع | عـ | ـعـ | ـع |
| Gheen | غ | غـ | ـغـ | ـغ |
| Faa | ف | فـ | ـفـ | ـف |
| Qaaf | ق | قـ | ـقـ | ـق |
| Laam | ل | لـ | ـلـ | ـل |
| Meem | م | مـ | ـمـ | ـم |
| Noon | ن | نـ | ـنـ | ـن |
| Haa | ه | هـ | ـهـ | ـه |
| Wow | و | و | ـو | ـو |
| Yaa | ي | يـ | ـيـ | ـي |

From DB we built, we conclude that the Arabic characters could be categorized into three main categories depends on their width. The first category includes small width letters; examples are "ا, لـ, ـلـ, ة, ه, ـا". The second is medium width category; examples are "ف, فـ, مـ, ـعـ". The third is the large width category with curves; examples are "ع, ر, ز, و, ج, ق, ف, ج".

## 4 EXPERIMENT RESULTS AND DISCUSSION

In this paper, we examined the character recognition using three different words that represent the three character width we defined. The words are: "شمس, جامعة, زرع". In our database, we focused on the letters of those words:

1. The First word "شمس": (8 images for sheen, 6 for meem, and 7 for seen).
2. The Second word "جامعة": (10 images for jeem and alif, 5 for aeen, and 10 for taa).
3. The last word "زرع": (4 for zaa and raa).

The first word "شمس" has 3 letters and they have the same segment width and all letters are connected with different positions. The word "جامعة" has 5 letters with different segment width for each letter and there is a space between "ﻟ" and "مـ" while the letter connections are variable. The third word is "زرع" has 3 letters. The letters are isolated and the two characters ر and ز are differentiated by only dot. Sometimes the letters overlapped each other for HW. Table 2 shows the recognition results of the segment of the three words. The segment width is 35 pixels.

TABLE 2
RESULTS OF CHARACTER RECOGNITION FOR THREE WORDS WITH FIXED SEGMENT WIDTH (35 PIXELS).

From the above table, we have recognition percentage of 25% with very high false recognition rate FAR. Applying our algorithm, the results shown in table 3 ensure increasing the recognition percentage to 55% with zero FAR.

TABLE 3
RESULTS OF THE CHARACTER RECOGNITION OF THE THREE WORDS WITH 1.3 SCALE FACTOR (55 PIXELS).

The results shown in table 3 are at scale factor of 1.3 of the base segmentation width. We have conducted many experiments to test the recognition percentage versus the scale factor used with the segment width so that it fulfills the highest recognition rate. The recognition percentage reaches 81% at scale factor 1.6 of the base width.

## 5 CONCLUSION

Building our own character DB with different character position shapes (Isolated, Initial, Medial, and Final), the authors run Arabic HW character recognition which coded using MATLAB to test three Arabic words cut from a sentence. These three Arabic words are best representing different character width and positions. We conclude that using fixed segment width gives low recognition percentage and high FAR. Resizing the segment width for the unrecognized character to adapt the different character width fulfill better recognition percentage of 81% and zero FAR.

For future work it is required to test our algorithm with more words and also tune the segment width so as to increase the recognition rate.

**Dr Manal Abdullah** received her PhD at computers and systems engineering, Faculty of engineering, Ain Shams University, Cairo, Egypt, 2002. She has experience in industrial computer networks and WSNs. Her research interest includes computer networks, performance evaluation, pattern recognition, WBANs, and bioinformatics. Currently she is assistant professor, Faculty of Computing and Information Technology FCIT, King Abdulaziz University, KAU, KSA.

**Lulwah M. Al-Harigy** is a demontartor in deanship of technology education and distance learning, King Abdulaziz University, KAU, KAS. She works as an instructional designer. She has a bachelor degree in computer science, Faculty of Computing and Information Technology FCIT, KAU. She is a master student at FCIT, KAU.

**Hanadi H. Al-Fraidi** is a demontartor in deanship of technology education and distance learning, King Abdulaziz University, KAU, KAS. She works as an instructional designer. She has a bachelor degree in computer science, Faculty of Computing and Information Technology FCIT, King Abdulaziz University. She is a master student at University of Ottawa, Ottawa, Canada.